**Autonomous on-Demand Shuttles for First Mile-Last Mile Connectivity: Design, Optimization, and Impact Assessment**

**Sudipta Roy**
Department of Civil, Environmental, and Construction Engineering
University of Central Florida, Orlando, Florida, United States, 32816
Email: sudipta.roy@ucf.edu
ORCID: 0000-0002-9588-6013

**Gabriel Dadashev**
The Porter School of Environment and Earth Sciences
Tel-Aviv University, Ramat Aviv 6997801, Israel
Email: dadashev@tauex.tau.ac.il
ORCID: 0000-0003-0674-5647

**Lampros Yfantis**
Scientific Researcher
Aimsun, Barcelona 08007, Spain
Email: lampros.yfantis@aimsun.com
ORCID: 0000-0002-0437-0223

**Bat-hen Nahmias-Biran**
The Porter School of the Environment and Earth Sciences
The School of Social and Policy Studies
Tel Aviv University, Tel Aviv 69978, Israel
E-Mail: bathennb@tauex.tau.ac.il
ORCID: 0000-0002-3223-4894

**Samiul Hasan**
Department of Civil, Environmental, and Construction Engineering
University of Central Florida, Orlando, Florida, United States, 32816
Email: samiul.hasan@ucf.edu
ORCID: 0000-0002-5828-3352

**ABSTRACT**
The First-Mile Last-Mile (FMLM) connectivity is crucial for improving public transit accessibility and efficiency, particularly in sprawling suburban regions where traditional fixed-route transit systems are often inadequate. Autonomous on-Demand Shuttles (AODS) hold a promising option for FMLM connections due to their cost-effectiveness and improved safety features, thereby enhancing user convenience and reducing reliance on personal vehicles. A critical issue in AODS service design is the optimization of travel paths, for which realistic traffic network assignment combined with optimal routing offers a viable solution. In this study, we have designed an AODS controller that integrates a mesoscopic simulation-based dynamic traffic assignment model with a greedy insertion heuristics approach to optimize the travel routes of the shuttles. The controller also considers the charging infrastructure/strategies and the impact of the shuttles on regular traffic flow for routes and fleet-size planning. The controller is implemented in Aimsun traffic simulator considering Lake Nona in Orlando, Florida as a case study. We show that, under the present demand based on 1% of total trips as transit riders, a fleet of 3 autonomous shuttles can serve about 80% of FMLM trip requests on-demand basis with an average waiting time below 4 minutes. Additional power sources have significant effect on service quality as the inactive waiting time for charging would increase the fleet size. We also show that low-speed autonomous shuttles would have negligible impact on regular vehicle flow, making them suitable for suburban areas. These findings have important implications for sustainable urban planning and public transit operations.

## INTRODUCTION

High population and economic growths in the urban regions of the USA are leading to increased traffic congestion, environmental impacts, and crashes. To reduce traffic congestion and associated problems, it is important to increase the use of public transit services which constitute about 1% of the mode share in the USA (*1*). Fixed-route transit systems in these areas are not cost-effective and do not provide quality services on a regular basis (2). As a result, less people use transit services, preferring private modes for their travel needs. To increase the mode share of public transit, especially in many mid and low-density cities and suburbs, one effective strategy is to integrate efficient access and egress systems (3).

The First-Mile Last-Mile (FMLM) connector service refer to the transportation solutions that bridge the gap between a commuter's starting point or destination and the nearest transit hub, such as a train or bus station to ensure efficient and convenient access to the transit system. Such services are often integrated with fixed-route transit systems to enhance their flexibility using various shared or private modes (3). Several case studies on integrating FMLM connector services with transit systems concluded that adding these services expanded transit catchment areas, increased accessibility, and significantly improved transit mode share (4, 5).

Although there are many advantages of introducing FMLM connector services into public transit, their effectiveness depends on service availability, reliability, and cost (3, 6). Researchers are studying various FMLM connectivity options based on these parameters. Autonomous electric vehicles (AEV) have gained worldwide interest due to their cost-effectiveness, reduction in travel mile, alleviation of parking problem and environmental benefits (7, 8) despite facing some key deployment issues including congestion, charging issues, and public perceptions (9–11). Autonomous on-demand shuttles can be operated based on individual travel needs and offer a promising solution for solving FMLM connectivity problems. There have been 57 instances of autonomous shuttle deployments in the USA, with most of them operating in controlled areas and all of them are free to ride (12). But some critical issues need to be addressed before commercial scale deployment of the autonomous shuttles for FMLM connector services. One issue is to find an optimal charging strategy for electric shuttles to ensure cost and energy efficiency (13, 14). Another issue is to mitigate the impact of low operating speed of the shuttles on road capacity (15). Addressing these issues requires optimization in several key areas including choosing optimal service mode and fleet size, travel routes, and charging strategies.

Previous research studied the integration of AEV or shared AEV in public transit with various features to increase efficiency, but still some research gaps exist in this domain (16–21). Design and optimization of on-demand shuttle services for FMLM connections through network traffic assignment considering charging strategies remain a novel issue and this is the research goal of this study. The objectives of the study include: (i) to design an Autonomous on-Demand Shuttle (AODS) controller by integrating dynamic traffic assignment with insertion heuristics to find optimal shuttle routes; (ii) to analyze the charging strategies for the shuttle fleet; and (iii) to assess the impact of the movement of low-speed autonomous shuttles on roadway capacity.

This study holds significant implications for both transit operators and riders. For transit operators, it provides a framework for designing efficient, reliable, and cost-effective FMLM connector services. This can lead to increased ridership and higher operational efficiency. The combination of dynamic traffic assignment and insertion heuristics in this work will be useful to reduce travel time and hence increase service and operational efficiency. Incorporating charging constraints will be useful to optimize fleet size and determine charging infrastructure requirement. Lastly, traffic impact assessment will provide likely impacts of shuttle movement on traffic network operations and insights on transportation infrastructure requirement.

## LITERATURE REVIEW

Previous studies focused on integrating autonomous vehicles/shuttles in public transit systems to serve FMLM connections. The operations of autonomous vehicles and shuttles can be differentiated in two aspects. First, the seat capacity of the shuttles can be up to 12 (22, 23), whereas autonomous vehicles can have up to 4-5 passengers (19, 21). The difference in the capacity increases waiting time for the shuttle as well as operating cost, which requires separate operational plans. Second, the operational speed of shuttles (mostly 10-15 mph (22, 23)) is significantly lower than autonomous vehicles (generally operates at road speed limit); as a result, sometimes shuttle operations are limited to low-speed roads (*24*).

Previous studies have analyzed different types of transit services, including fixed and demand-responsive routing services. In the fixed service type, travel routes and stops remain fixed but for the demand-responsive system, the routing and the stops are determined completely based on real-time demand information (22). Studies (22), (25) have compared fixed and on-demand services and showed that the fixed routing service is superior for very high demand and in case of low demand the demand-responsive service performs well. (26) also concluded that demand-responsive transit should be preferred over fixed routing for low operational cost and better services for operating in low demand regions.

Previous studies mostly used agent-based modeling (ABM) for integrating autonomous vehicles/shuttles with public transit systems, with majority of them focused on autonomous vehicles only. ABM can be used to simulate complex behavioral dynamics of supply-demand interaction, stochastic nature of the operations, real-time adjustment, and related decision-making for vehicle fleet (27). Some of the relevant works on autonomous vehicle/shuttle integration to transit systems are presented in **Table 1**. All these studies used different tools, but ABM approach was followed which could accommodate the dynamic interactions between vehicle-rider efficiently. Despite these advancements, only two studies focused on charging strategies for the service operator (24, 28), highlighting a gap in understanding the operational challenges of integrating electric vehicles/shuttles in the service. Also, none of these studies have focused on the impacts of autonomous shuttles on regular traffic flow which will help find infrastructure changes to seamlessly integrate autonomous shuttles to existing road networks.

Optimizing travel routes is a major issue for on-demand shared vehicle movement as reducing travel time and cost is the key to offer an efficient service. Exact solution methods are the most accurate route optimization methods, but they are difficult to apply in complex routing problems (29). In contrast, heuristics-based optimization methods offer close to optimal solutions with much lower computation time (29). Different types of heuristics are applied for solving vehicle routing problems in autonomous on-demand vehicle deployment. Some studies focused only on the nearest vehicle search for computational efficiency (19, 30). But most studies used different variations of insertion heuristics for trip assignment and vehicle routing which provided efficient solutions (21, 22, 31). Only few studies integrated insertion heuristics with a dynamic traffic assignment model to incorporate dynamic path cost for choosing routes (30, 31) (**Table 1**).

**TABLE 1. Relevant studies on simulation-based AV deployment as FMLM connector service**

| Study | Real Network? | Service Type | Autonomous Vehicle/ Shuttle? | Considered dynamic path cost for route choice? (In case of On-demand service) | Vehicle-Rider Matching Algorithm (In case of On-demand service) | Considered Charging Strategy in Modeling? | Considered Impact Assessment on Regular Traffic? |
|---|---|---|---|---|---|---|---|
| **Scheltes et al., 2017** (28) | Yes | On-demand | Private Ride-hailing | No | Nearest option search | Limited Consideration | No |
| **Wen et al., 2018** (21) | Yes | On-demand | Shared Vehicle | No | Insertion heuristics | No | No |
| **Shen et al., 2018** (19) | Yes | On-demand | Shared Vehicle | No | Nearest option search | No | No |
| **Gurumurthy et al., 2020** (30) | Yes | On-demand | Shared Vehicle | Yes | Nearest option search | No | No |
| **Lau and Susilawati, 2021** (32) | Yes | On-demand | Shared Vehicle | No | Predefined routes | No | No |
| **Huang et al., 2022** (31) | Yes | On-demand | Shared Vehicle | Yes | Rule-based vehicle rider matching algorithm (like insertion heuristics) | No | No |
| **Rich et al., 2023** (22) | Yes | Both Fixed and On-demand | Shuttle | No | Insertion Heuristics | No | No |
| **Grahn et al., 2023**(24) | Yes | On-demand | Shuttle, Private Ride-hailing, Shared Vehicle | No | Insertion Heuristics | Limited Consideration | No |
| **Aalipour and Khani, 2024** (33) | Yes | On-demand | Private Ride-hailing | No | Linear Time-Delay | No | No |
| **Roy et al., 2024 (this study)** | Yes | On-demand | Shuttle | Yes | Insertion Heuristics | Yes | Yes |

From the above literature review, we identified several gaps which include: (i) lack of studies that analyzes autonomous shuttle operations after integrating dynamic traffic assignment and optimization for shuttle routing (ii) lack of studies assessing charging strategies; and (iii) lack of analysis on the impact of low-speed autonomous shuttles on roadway capacity. This study aims to fill these gaps by proposing a novel simulation-based framework for designing and optimizing an autonomous on-demand shuttle service for FMLM connections. The framework integrates agent-based demand models with dynamic traffic simulation to mimic real-world demand-supply interactions. It also includes evaluating charging strategies which was considered in a limited way in only two prior studies. Finally, it evaluates the impact of shuttle operations on roadway capacity to assess the feasibility of such operations.

## METHODOLOGY

In this study, we analyze the operations of autonomous on-demand shuttle (AODS) service to enhance FMLM connectivity in a sub-urban context. This methodology employs a state-of-the-art traffic simulator Aimsun and Aimsun Ride API for managing the vehicle-passenger interaction and dynamic vehicle routing (34, 35). We have adopted an event-based modeling approach that allows for real-time decision-making, adapting to the dynamic nature of passenger requests and traffic conditions. This system optimizes shuttle routes to minimize waiting times and travel distances, ensuring efficient shuttle operations.

### Study Area

The case study of this research is implemented over Lake Nona which a fast-growing community located in Orlando, Florida. This area has diverse residential, commercial, and recreational spaces, creating a dynamic and attractive environment for residents and visitors. The core residential area spans approximately 3 square miles and includes around 2500 residential parcels and other 500 commercial or recreational parcels (**Figure 1**) (36). The area has a variety of points of interest which attract a high number of daily commuters and visitors.

Currently, Orlando's public transit agency, Lynx, operates three routes serving Lake Nona, with regular intervals at a few transit stops throughout the day. However, the suburban nature of the area and the spread-out residential and commercial locations present challenges for residents and visitors trying to access these transit stops. Implementing FMLM solutions can significantly enhance the connectivity of transit users by facilitating easy and quick access to transit stops. The suburban area type with potential transit users makes Lake Nona ideal for testing AODS deployment. Additionally, most roads have a speed limit of 25 mph, making it safer for low-speed autonomous shuttle movements. Lake Nona currently has an autonomous shuttle service which is operated (on limited times and routes) by a private company Beep (37).

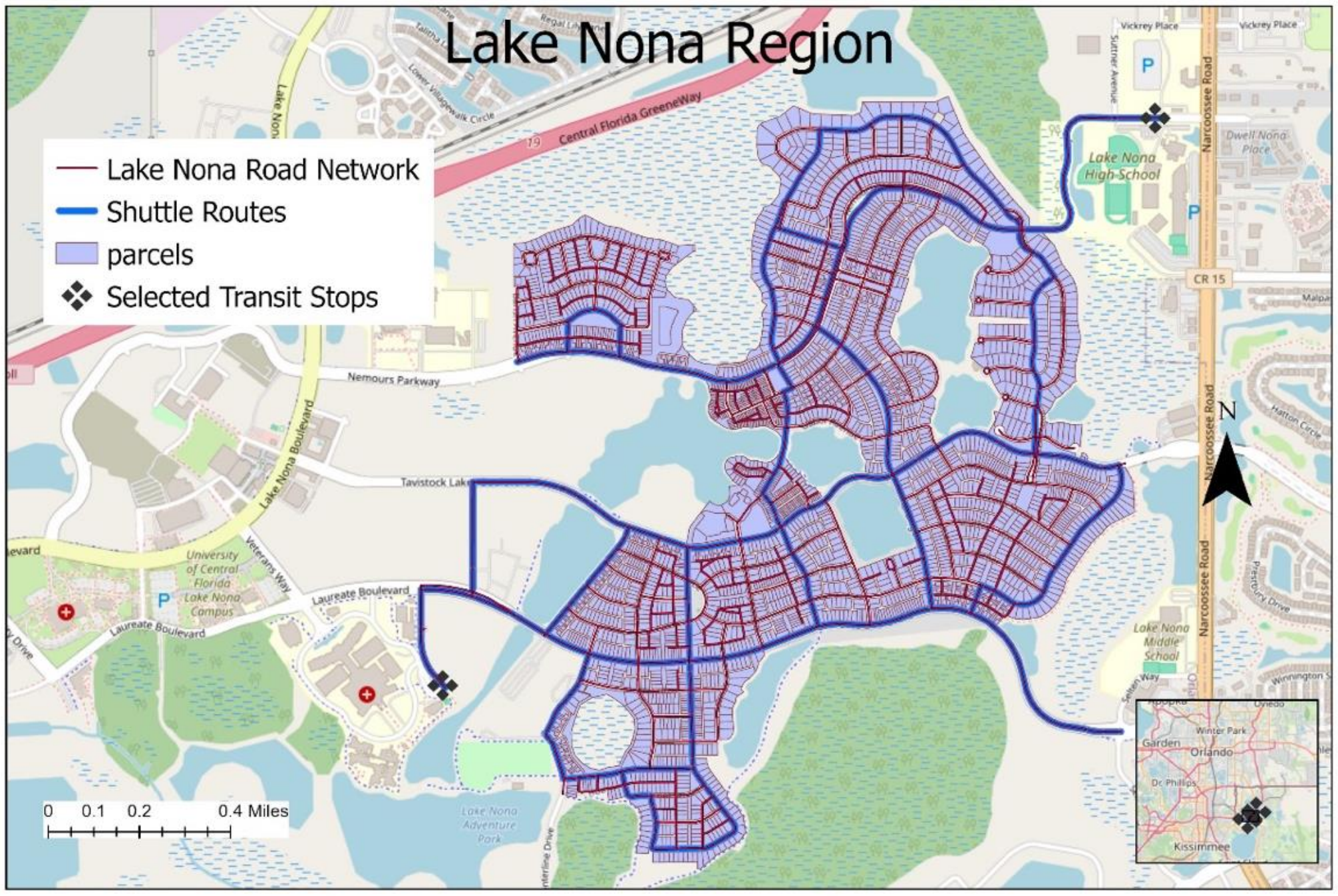

**Figure 1. Study area at Lake Nona Region**

**Autonomous on-Demand Shuttle (AODS) Service**

The Autonomous on-Demand Shuttle (AODS) service will feature autonomous electric buses operating at 15 mph, with a passenger capacity of 8. These shuttles will travel on fixed road sections within the study area (**Figure 1**). The length of the shuttle route is 21.2 miles (one-way distance). Proposed fixed routes offer two advantages: (i) they enhance safety by simplifying the autonomous navigation process, as the shuttles follow known paths; and (ii) they allow better integration with existing traffic systems and improve service reliability. This shuttle service will be designed to operate on a stop-to-stop system, with dedicated stops strategically placed throughout the study area (will be discussed in the next section). This approach will enhance the operational efficiency of the shuttle and better accessibility for passengers. Two transit-stops are selected for the transit connectivity (**Figure 1**).

**Placement of Shuttle-stops**

Shuttle-stops are placed using a K-means clustering algorithm. The main objective is to minimize the number of shuttle stops while providing better accessibility to users, ensuring minimal walking time from/to their parcel origins and destinations. Initially, a network is created using Python's NetworkX package (38). K-means clustering is then applied to the unit parcel coordinates to determine cluster centers, where each cluster center acts as a shuttle-stop for the nearby parcels. These cluster centers are matched with the nodes in the road sections where the shuttle is operating.

Assuming a standard walking speed of 3 mph, a coverage matrix is produced to show the percentage of parcels covered under various scenarios. Variables include the number of shuttle-stops (ranging from 8 to 24) and the maximum walking time (3 to 7 minutes). The results suggest that with 15 shuttle-stops and a maximum walking time of 6 minutes, 96% of the parcels can be covered. This configuration is determined to be the optimum number of shuttle-stops for the study area.

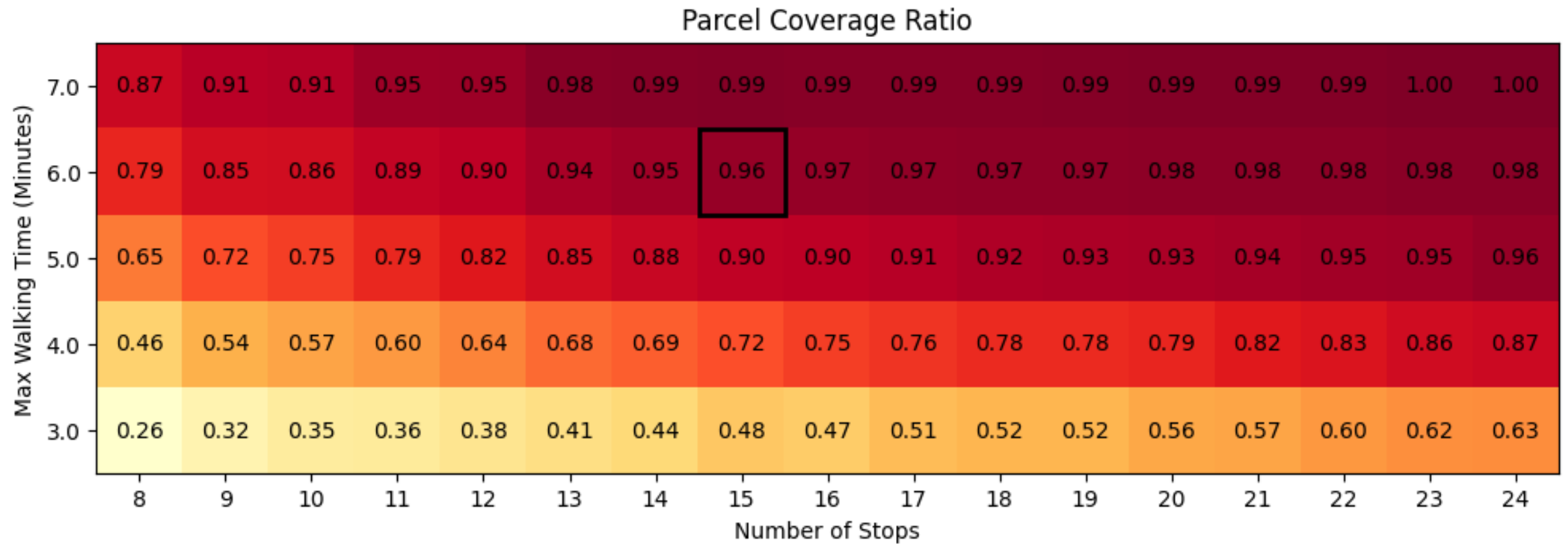

**Figure 2. Parcel coverage ratio matrix**

The centroids of the shuttle and transit stops are determined using k-mean clustering to identify a central location for the garage and charging points. This centroid location is then adjusted to find a suitable nearby place that can serve as a vehicle storage and charging area without causing inconvenience to other road users. The centroid location, the adjusted vehicle garage and charging points location, along with the shuttle and transit-stops are shown in **Figure 3**.

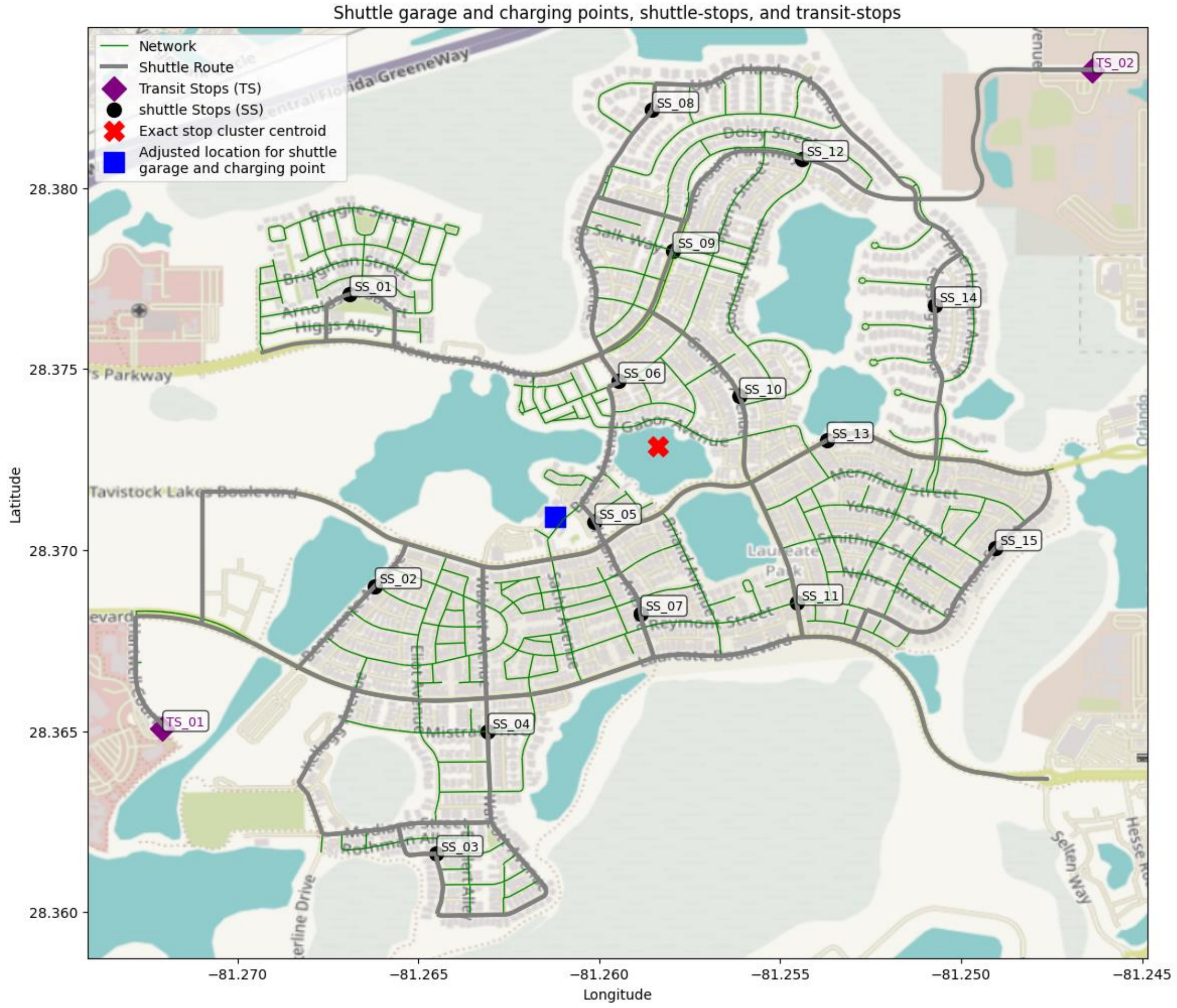

**Figure 3. Locations of shuttle garage and charging points, shuttle stops, and transit stops.**

**Preparation of Simulator and Traffic Demand**

The simulation network in Aimsun is prepared by first importing the raw network from OpenStreetMaps (39). The raw network is manually modified to ensure an accurate representation of the real-world scenario, including detailed road layouts, intersections, and other essential infrastructural elements necessary for precise traffic simulation. The simulation duration will be 13 hours (6 AM to 7 PM).

Within this network, seven internal centroids are created to generate internal traffic demand, strategically placed to capture travel demands originating within the study area. Additionally, seven external centroids are positioned to capture the demand from major roadways surrounding the study area ensuring the integration of external-to-external, external-to-internal, and internal-to-external traffic demand into the simulation.

The initial demand data for 2019 is provided by Teralytics Studio which collects mobility data using a combination of different sources, i.e., anonymized GPS location data from smartphones, dynamic traffic volume count data of road detectors and public-reference data including census and employment information (40). This data is subsequently calibrated to reflect 2024 conditions using OD adjustment and OD departure adjustment procedures in Aimsun, utilizing Regional Integrated Transportation Information System (RITIS) road detector data available for SR417. Six detectors are used for calibration and other four detectors are used for validation (41, 42). This calibration process adjusts and validates the simulated travel patterns with a GEH score of 3.94, ensuring an accurate representation of current traffic conditions. **Figure 4** shows the calibrated demand.

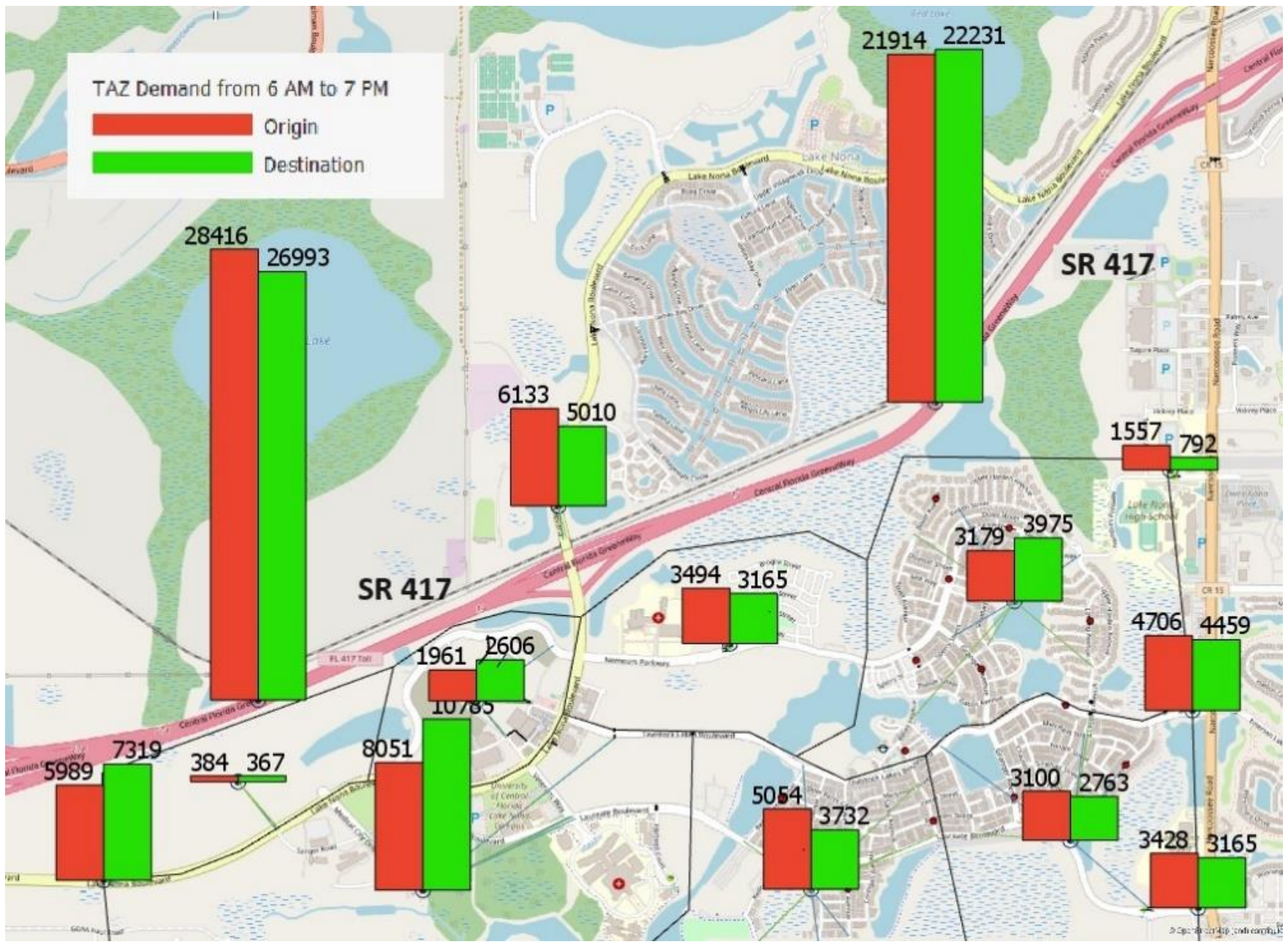

**Figure 4. Calibrated O-D demand in the study area**

**Simulation Experiments Set-up**

In this study, a mesoscopic simulation approach is employed for dynamic traffic assignment due to low running time required to simulate each scenario. As we are not concerned about the microscopic vehicle interaction parameters here, mesoscopic simulation can provide adequate level of details in this context. The simulation process begins with performing Dynamic User Equilibrium (DUE) to generate optimal paths between each origin-destination (OD) pair in 15-minute intervals. Aimsun uses a gradient-based model to find equilibrium conditions (43).

Two types of vehicles are simulated here: cars and autonomous shuttles. Car movements are generated by an OD matrix, reflecting typical traffic flows. For shuttle movements, the fleet size of shuttles is controlled by the event-based external shuttle controller, and the shuttle's path is determined through a combination of dynamic traffic assignment within the simulation and the controller's logic.

**Trip Request Generation Procedure**

The process of generating trip requests for the FMLM services begins with estimating the total external-to-internal and internal-to-external trip demand in the area. The calibrated OD matrix for car (described in the previous section) is used to calculate the total incoming and outgoing trips attracted to and generated from the area in each hour.

From the National Household Travel Survey (NHTS) 2017 and 2022 data (1), we have found that approximately 1% of trips are associated with public transit in the South-Atlantic metropolitan statistical area (MSA) in cities with a population of more than 1 million without heavy rail. Since these characteristics match with our study area, we have used this ratio to estimate the first mile (FM) and last mile (LM) trip requests. Thus, we have assumed that approximately 1% of the total outgoing trips would use the shuttle service as an FM trip connector and the same proportion would use the shuttle service as the LM trip connector. Additionally, we assumed that the distance of FM and LM trip lengths to use shuttles should be a minimum of 1 mile. To spatially allocate FM and LM trips into different transit and shuttle stops, a random assignment process is used within each time interval. First, the origin-destination pairs are analyzed for the whole area for which the trip distances would be a minimum of 1 mile. Then, the calculated number of FM and LM trips are randomly assigned to the stops. The total number of FM and LM trips are calculated as 125 and 123, respectively for the full simulation period of 13 hours. We have also calculated the demand for a more futuristic scenario where 2% of the total trips might be associated with transit. In that scenario, the total calculated FM and LM trips for the same simulation duration are respectively 226 and 223. The distribution of FM and LM trip requests across the simulation period are shown in **Figure 5a and 5b**.

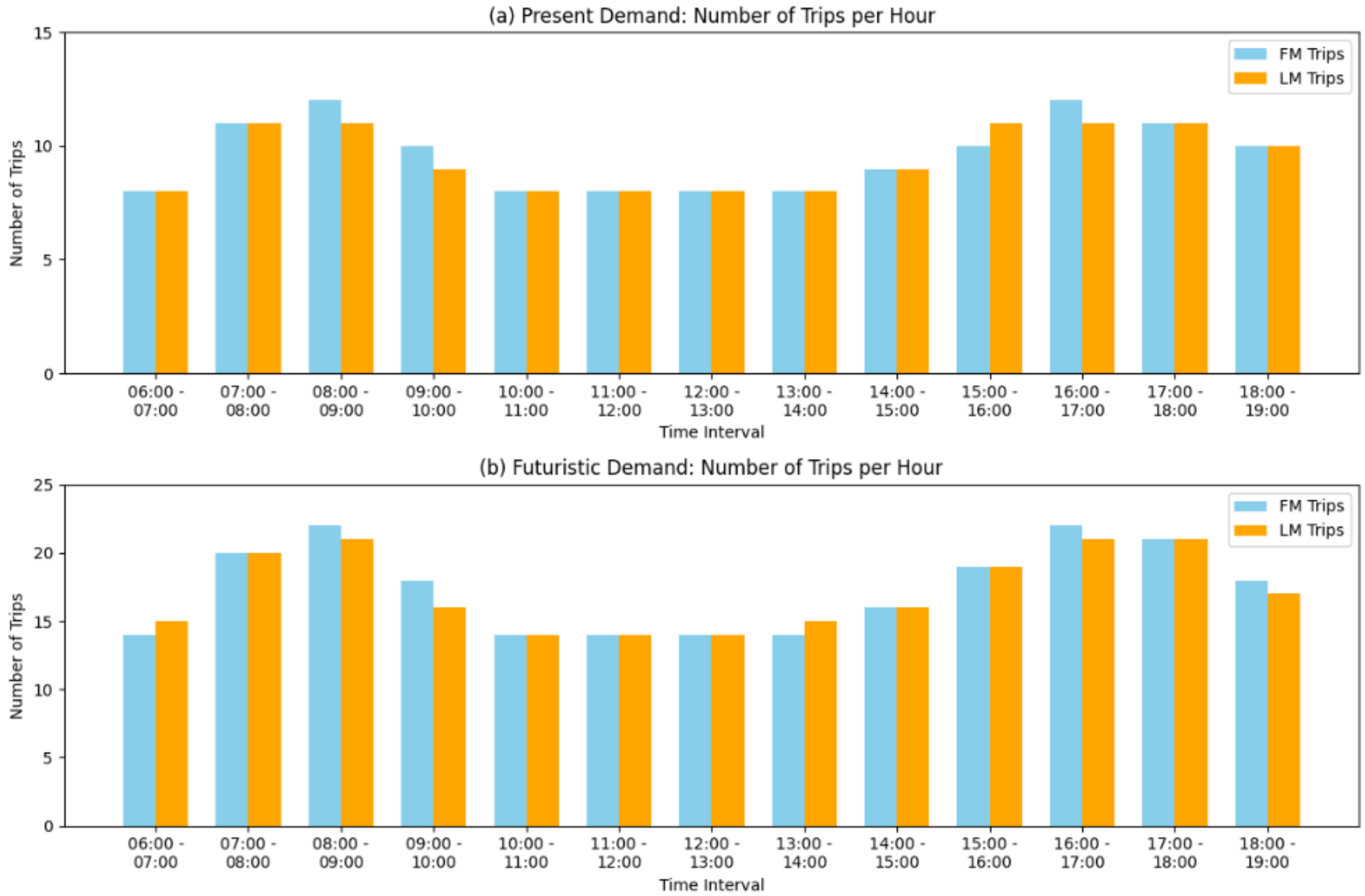

**Figure 5. FM and LM trip requests for shuttle: (a) present; (b) futuristic demand**

**Charging Attributes for the Electric Shuttles**

We have assumed that the battery specifications of a Navya autonomous shuttle, which is equipped with a battery pack, has a theoretical full charging capacity of 33 kWh (23). For practical purposes, the charging capacity is set to 30 kWh (90% of the full capacity). With an energy efficiency coefficient of 0.80 applied, the shuttle has a per mile charge requirement of approximately 0.30 kWh/mile and a per minute charge requirement of approximately 0.08 kWh/minute when idle.

To support fast and efficient recharging, the DC Level 3 chargers can be a perfect option which can recharge EV batteries with more than 50 kW capacity (44). We assume that this type of chargers with a capacity of 50 kW will be used to charge the shuttle. This high-capacity charger can recharge the shuttle's battery to its full capacity in 45 minutes (applying 0.80 as capacity factor), assuming a linear relationship between charging time and battery capacity. We have assumed some battery condition check rules here. The controller would check for recharging options after the charge level becomes less than 15 kWh and stops operation after 5 kWh at idle state if it still doesn't get any recharging option. At that point, it waits and checks for recharging options at 1-minute frequency.

**Controller Design**

We have designed a controller in Aimsun Ride for running the shuttles. Aimsun Ride is a plug-in for Aimsun simulator; it is designed to model the operations of on-demand mobility services, fleet management, and subsequent shuttle-rider interaction within simulation environments. The controller is coded in Python and run from an external environment interacting with the Aimsun simulator. **Figure 6** shows the controller processes which are divided into two parts: an offer module and a trip module.

In the offer module, when the simulation starts, every shuttle is considered as active and ready to receive any trip request. When a ride request arrives, the system searches for active shuttles and checks the available number of seats of each shuttle. If any shuttle has the required number of seats, it employs the shuttle-rider matching algorithm (details given in the next section) to retrieve feasible schedules to serve that request. If no active vehicle or feasible schedule is found, the request is added to failed request list. The controller finds the least costly path after running the mesoscopic dynamic traffic assignment (DTA) algorithm and thus optimizes traveling routes of the shuttles. The schedule with the least travel time is offered for the trip when the trip module starts.

The trip module controls the events of the trip (e.g., start traveling to origin, pickup passenger, start traveling to destination, and drop off) and each event initiates a particular state for the shuttle. At the change of each state (e.g., traveling to origin, at pickup, pickup done, traveling to destination, at drop off, drop off done, and Idle), the occupancy of the shuttle and the battery condition rules are checked, and the shuttle status is updated accordingly. The controller has two main component including real-time feasible schedule generation algorithm and the charging algorithm, which are discussed in the following sub-sections.

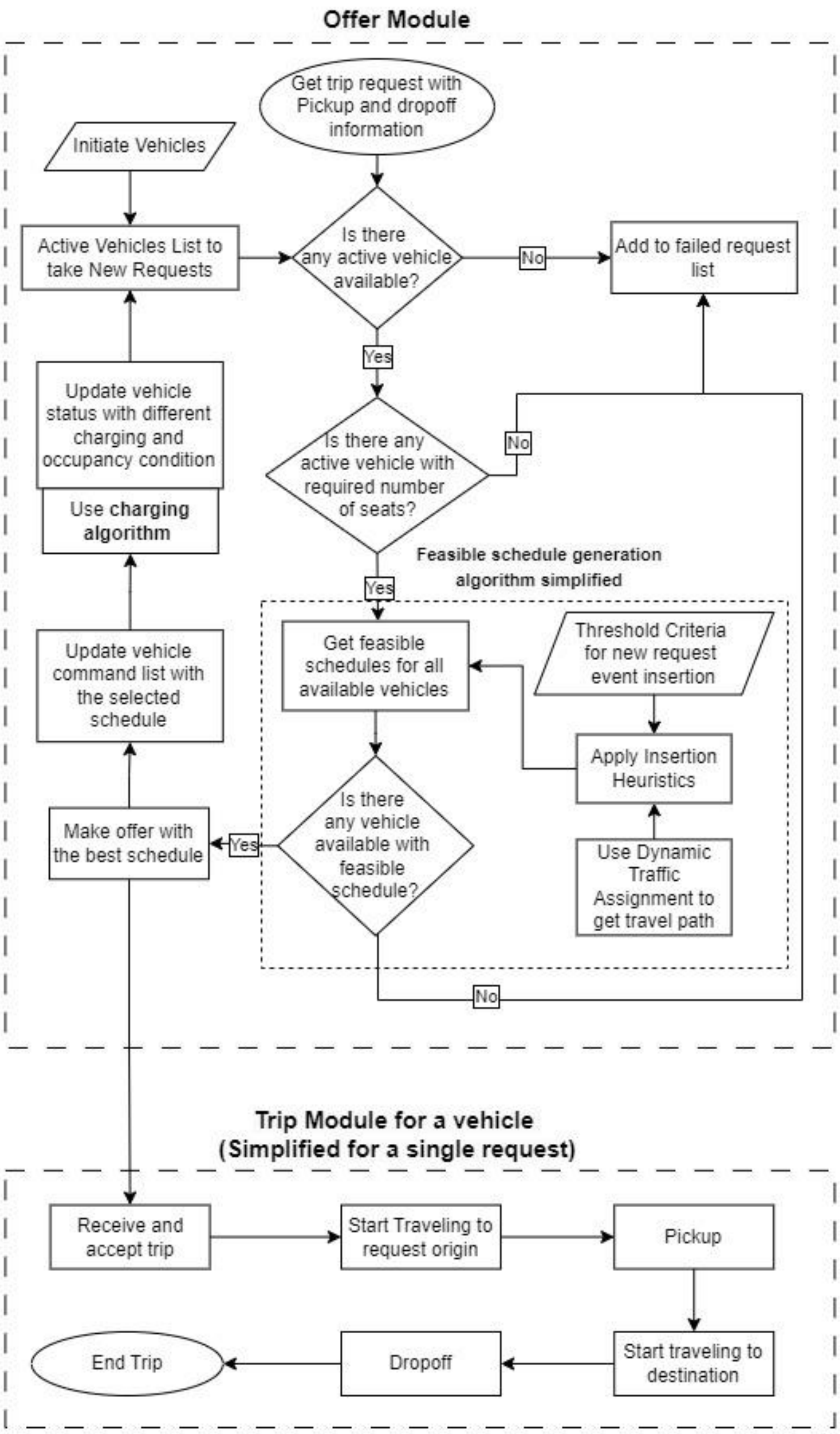

**Figure 6. The shuttle controller design for FMLM connector service** (inputs are in parallelograms, start/end of processes are in ovals, decision-making steps are in diamonds, and the operation steps are in rectangles.)

*Real-time Feasible Schedule Generation Algorithm*

The controller follows a structured approach to generate the list of feasible schedules for shuttles to serve the trip requests (**Figure 7**). It uses a greedy insertion heuristics optimization algorithm which is a state-of-the-art algorithm for vehicle routing and scheduling (29). This is considered as a greedy algorithm because it makes locally optimal choices at each step to achieve immediate benefits, aiming for the global optimum. This approach does not account for how the insertion of the current request might affect the accommodation of future requests and does not involve iterating back to reconsider previous decisions. Despite these limitations, similar methodology has been shown to efficiently serve trip requests in previous studies (22, 45).

After finding that an active shuttle has the requested number of seats in the offer module, the shuttle-passenger matching algorithm proceeds with the following steps. First, it checks if the vehicle is empty. If it is empty, it receives the travel path and time for that trip using DTA algorithm. The travel schedule is created directly by inserting the pickup and drop off commands sequentially, and then it checks if this schedule meets the offer threshold criteria of maximum waiting time. However, if the shuttle already has a running schedule, the process involves multiple checks.

The pickup and drop off commands are sequentially inserted at different points in the existing schedule to create a potential new schedule. Then for each new potential schedule, the travel path and time is calculated using DTA algorithm. Each potential schedule undergoes three checks: (i) at first, it checks if the waiting time and travel time increment for the new trip meet the defined thresholds; (ii) if these conditions are met, it checks whether the new schedule will significantly impact the travel time for ongoing trips in the shuttle; (iii) finally, it checks if the new schedule will affect the passengers who have booked but not yet been picked up.

Once a potential new schedule passes all these checks, it is added to the feasible schedule list. The best schedule (with respect to travel time) is selected, assigned to the shuttle, and the offer is generated.

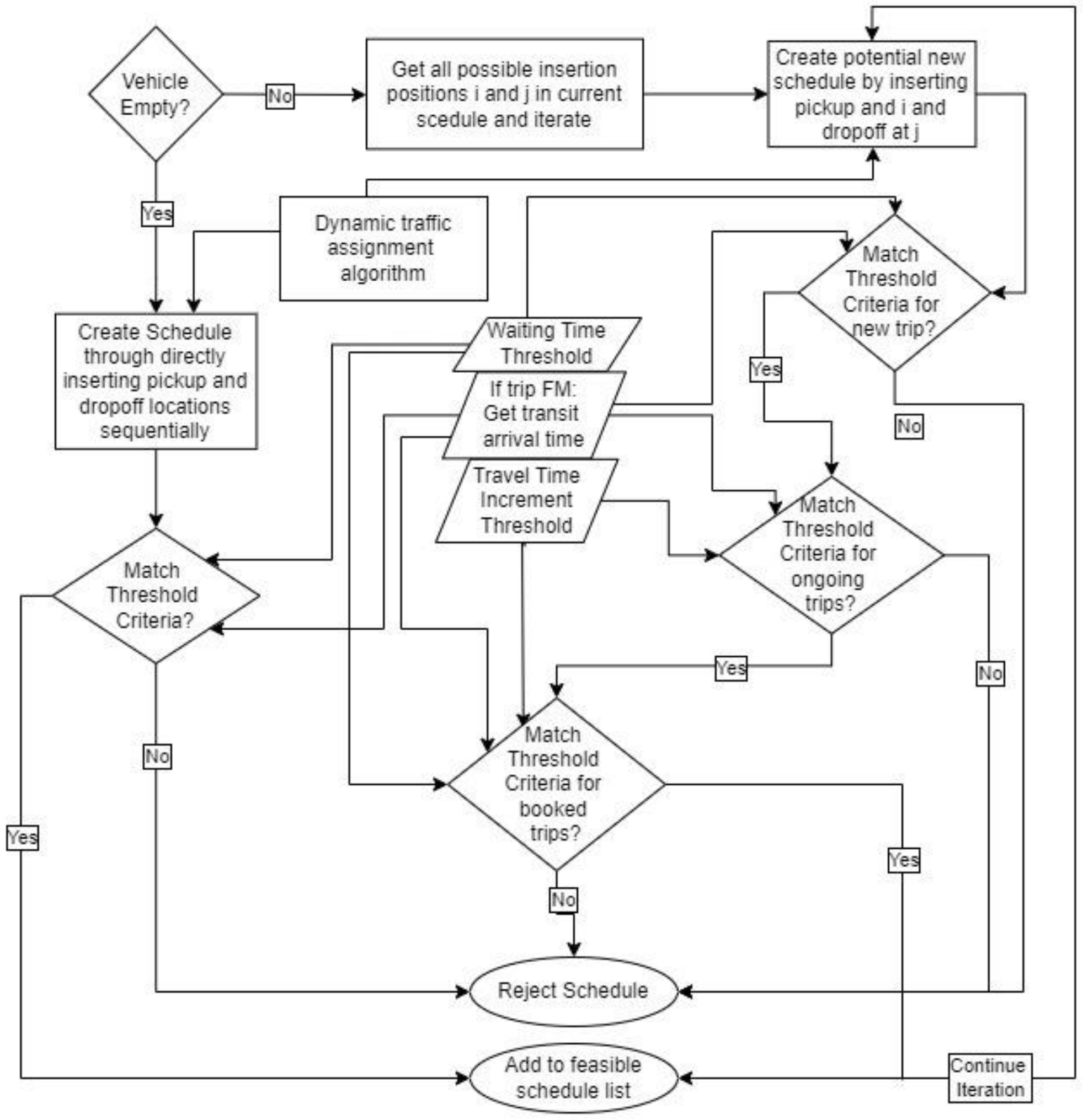

**Figure 7. The real-time feasible schedule generation algorithm for all available vehicles**

*Shuttle Recharging Algorithm*

The objective of the shuttle recharging algorithm is to maintain sufficient battery levels to provide reliable service while optimizing the use of available charging points (see **Figure 8**). The process begins by detecting any change in the state of a shuttle by calculating the distance traveled, time spent traveling, and energy consumed since the last update to check the current battery status. If at any point of operation, current battery status falls below critical limit, then the vehicle's active status is turned off, which means it won't accept any new request. If a shuttle is in the 'Idle' state, the system evaluates if recharging is necessary by checking the threshold. If recharging is required, the shuttle is added to the charging waiting list if it is not already on the list. The algorithm then checks if the vehicle is at the priority positions of the waiting list and if any charging points are available. If both conditions are met, the shuttle is directed to reposition for charging, and its active status is turned off.

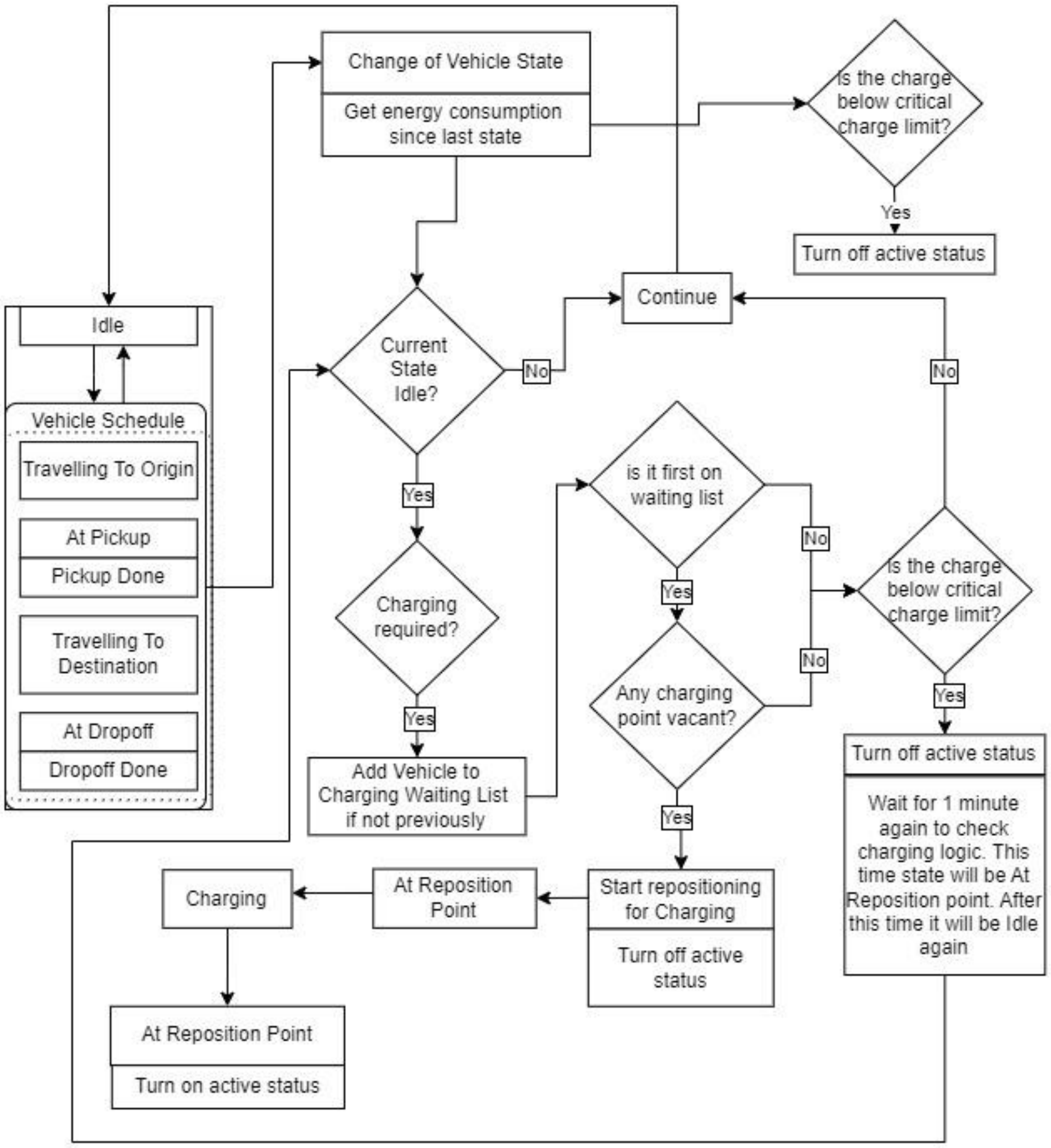

**Figure 8. The shuttle recharging algorithm**

Once the shuttle transitions to the Repositioning state and reaches the assigned charging point, it begins the charging process. After the charging is complete, the vehicle's battery level is restored to full charge, its active status is turned back on, indicating it is ready for service again and it remains on Idle condition. In cases where a vehicle's charge level falls below a critical limit and it is not the first in the waiting list or there are no vacant charging points, the vehicle's active status is turned off, and the system waits for a short duration before rechecking the charging logic. This ensures efficient utilization of available charging infrastructure, prioritizes vehicles based on their energy status.

### Controller Configurations Variations in Scenarios

We have created a series of simulation scenarios by varying different configurations as listed below:

- **Shuttle fleet size**: The number of shuttles operating in the system varying from 2 to 6.
- **Maximum waiting time**: The maximum waiting time for passengers before the shuttle arrives are set as 6, 8, or 10 minutes.

- **Travel time increment coefficient threshold**: The acceptable increase in travel time of existing rides to include new rides in shuttle schedule. The threshold is set at 0.5 or 1 (no units).
- **Charging points:** The number of available charging points for the shuttles considering 1 or 2 charging points.

We generated a set of scenarios by making 60 unique combinations with these parameter variations for both present and futuristic demands and we made 4 random instances for both demand profiles. As a result, out total generated simulation scenarios are 60*2*4 = 480.

## RESULTS

In this section, we will analyze the output parameters from the scenarios. We have divided the output parameters in four categories: rider-based, vehicle-based, energy consumption-based, and traffic impact-based parameters. We examine these parameters across different fleet sizes and maximum waiting times. For each parameter, the analysis includes mean and standard deviation values, aggregated across different random iterations of trip request generation.

### Rider-based Parameters

The rider-based parameters include accepted requests to total requests ratio, median total trip time to walking time ratio, and median waiting time (**Figure 9**). The result suggests that increasing the fleet size consistently improves all rider-based parameters for both present and future demand scenarios with approximately 250 and 450 generated requests, respectively.

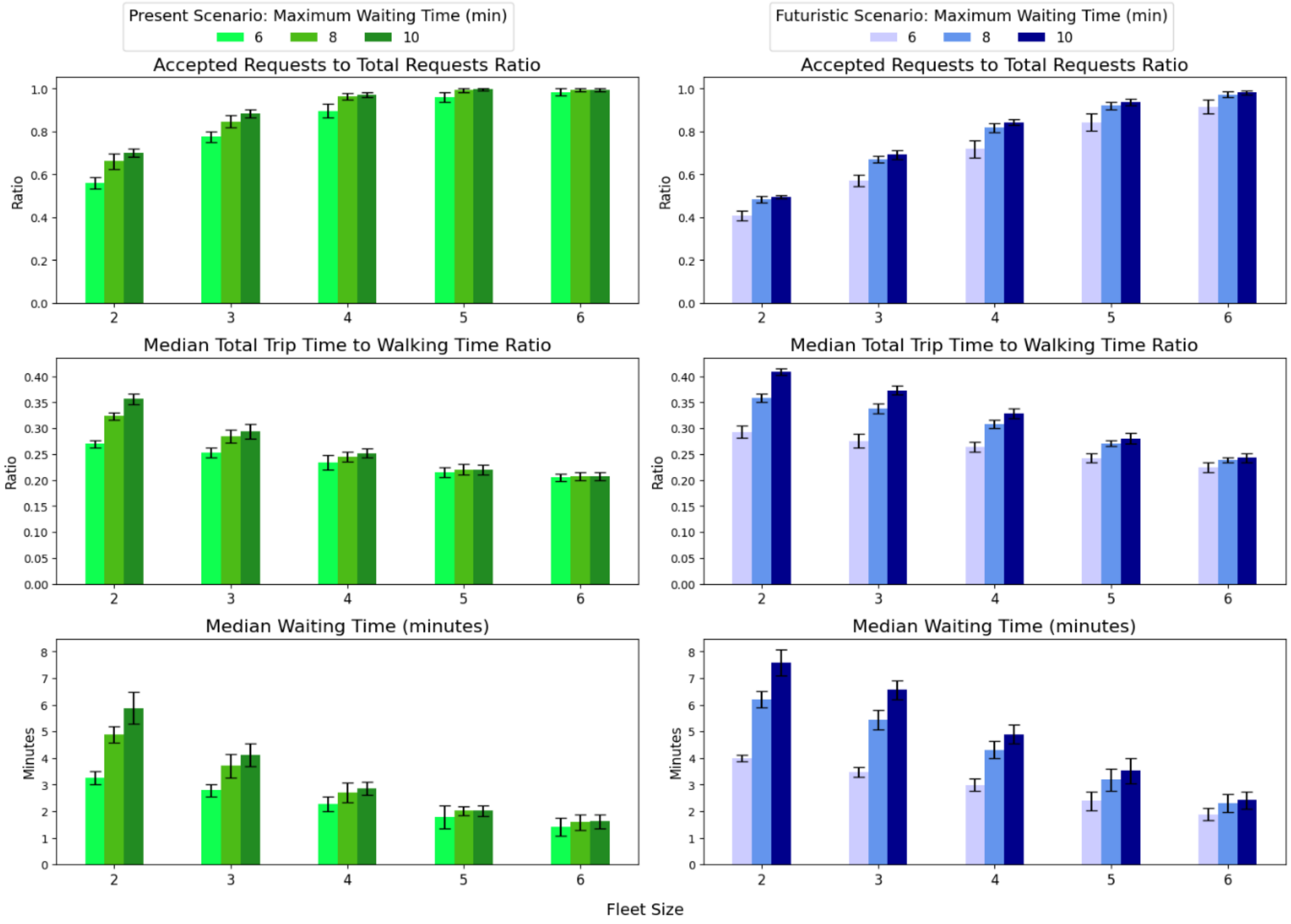

**Figure 9. Rider-based parameters**

The accepted requests to total requests ratio gradually improves with increasing fleet size and maximum waiting time and there is a trade-off between the fleet size and maximum waiting time limit. The result suggests that about 80% of requests can be served with a relatively smaller fleet of 3 shuttles in present scenario with a maximum waiting time limit of 8 minutes (while the actual median waiting time is less than 4 minutes). The future demand scenario would require 4 shuttles to get same accepted requests to total requests ratio maintaining almost similar waiting time. The median of total trip time to walk time ratio suggests that if we choose even the lowest fleet size of 2 shuttles and 10 minutes maximum waiting time limit, the total trip time (waiting time plus in-vehicle travel time) never cross 40% limit, which means that the shuttle performance is at least (100/40 = ) 2.5 times better than the walking alternative with respect to travel time. Although the total trip time to walking time stabilizes around 20% which implies an operational saturation point where the additional shuttles or reduced waiting times does not significantly improve time.

**Vehicle-based Parameters**

The vehicle-based parameters include total travel distance, empty vehicle travel to total travel ratio, idle time to total active time ratio, and shuttle capacity utilization ratio for each experiment (**Figure 10**).

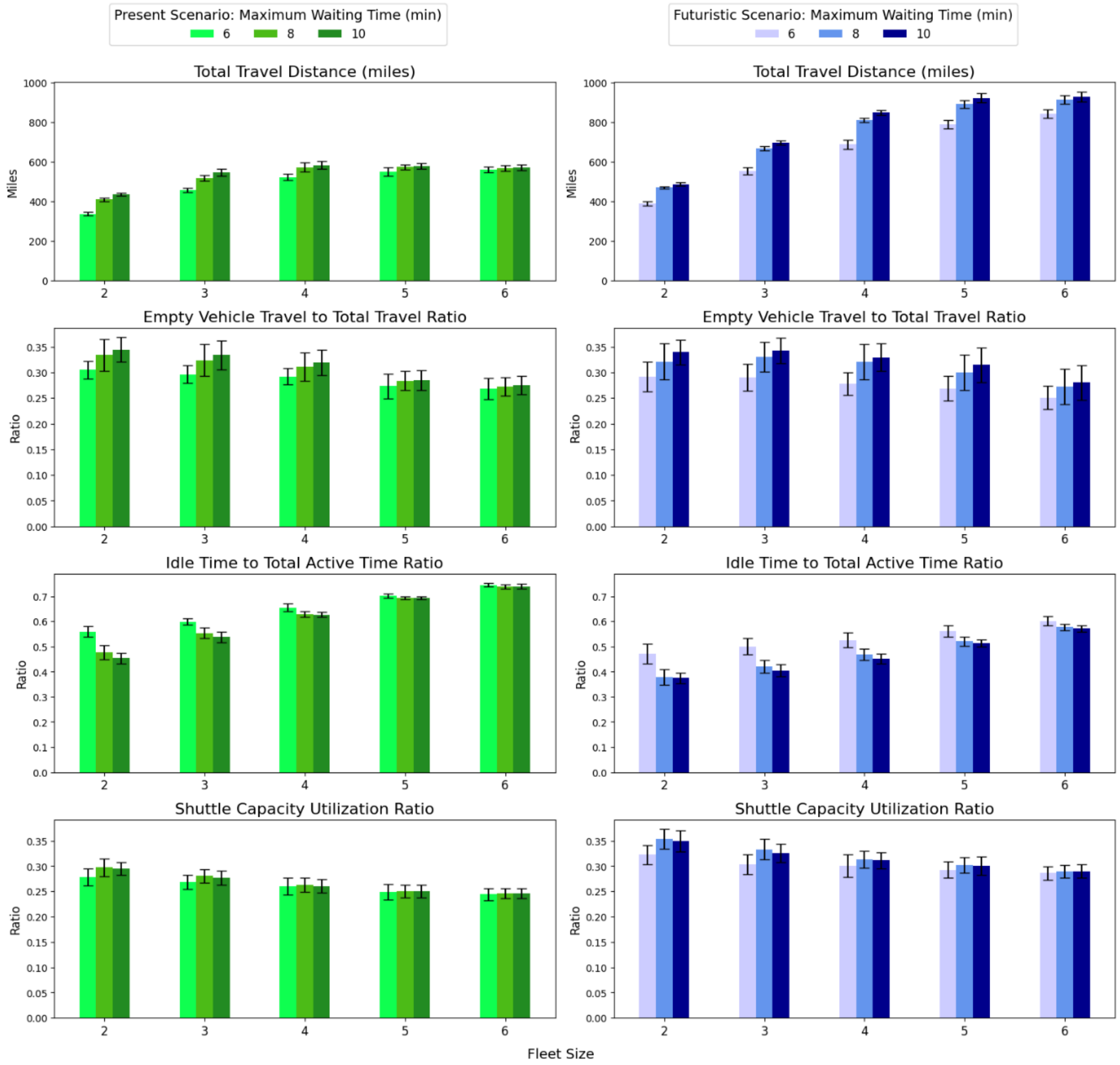

**Figure 10. Vehicle-based parameters**

The total distance traveled by all the shuttles of a fleet (**Figure 10**) increases with fleet size. However, beyond a particular fleet size, additional shuttles do not significantly increase the total travel distance, indicating an optimal fleet size for operational efficiency.

The empty vehicle travel (the travelling distance without any passenger) to total travel ratio remains relatively stable across different fleet sizes which is around 30% for both demand scenarios. This indicates regardless of the fleet size, that percentage of the total vehicle traveled will be without any passenger. But this ratio has a positive correlation with the maximum waiting time limit, as with increasing waiting time, the shuttles travel longer distances to pick up the passengers.

The idle time to total active time (total idle time plus total travelling time) ratio for the shuttles increases significantly with the fleet size and is negatively correlated with waiting time threshold. Larger fleets result in higher idle times due to more shuttles being available for fewer demands.

The shuttle capacity utilization ratio is the sum of (shuttle occupancy * durations of the occupancy) for all shuttles in the fleet divided by (total shuttle capacity * total travelling time for all shuttles in the fleet). Despite the fleet size increase, there is not much effect on the shuttle capacity utilization ratio, it remains close to 25% on average, which suggests that on average 2 out of 8 seats remain occupied. For the futuristic condition, this average just slightly increases to 30%.

### Energy Consumption-based Parameters

**Figure 11** presents the total energy consumption parameter for both demand scenarios. Energy consumption for shuttles increases almost linearly with fleet size. This suggests that the additional idle time associated with a larger fleet size does not significantly impact energy consumption, as shuttles continue to consume energy during idle periods, albeit at a lower rate. The energy consumption slightly increases in the futuristic demand scenario suggesting that the substantial increase in trip demand in the futuristic scenario has a little effect on overall energy consumption.

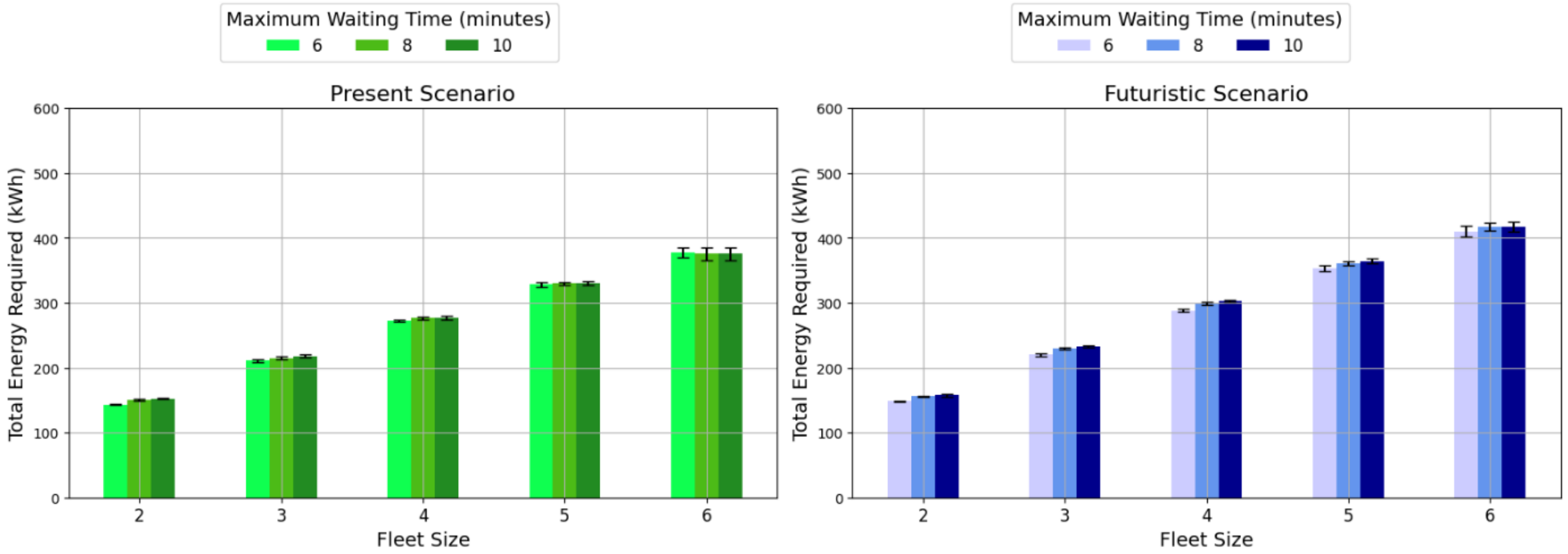

**Figure 11. Total energy consumption**

The charging capacity utilization ratio is ratio of the total charging time for all shuttles in a fleet to total available charging time (**Figure 12)**. It suggests that using only 1 fast Level-3 DC charging units can be enough even for a fleet of 6 shuttles, where only about 60% of the time the charging point will be occupied. However, with increasing fleet size, the likelihood of higher waiting times for charging also rises; installing two charging units would significantly reduce waiting times for charging. The parameter inactive charging waiting time to total charging time ratio is introduced to analyse the waiting scenario. It is observed that specially for the scenarios with shuttle fleet size = 5 or 6, the mean of inactive waiting time to total charging time ratio is significantly high. This also indicates the need of 2 charging units for serving higher demand.

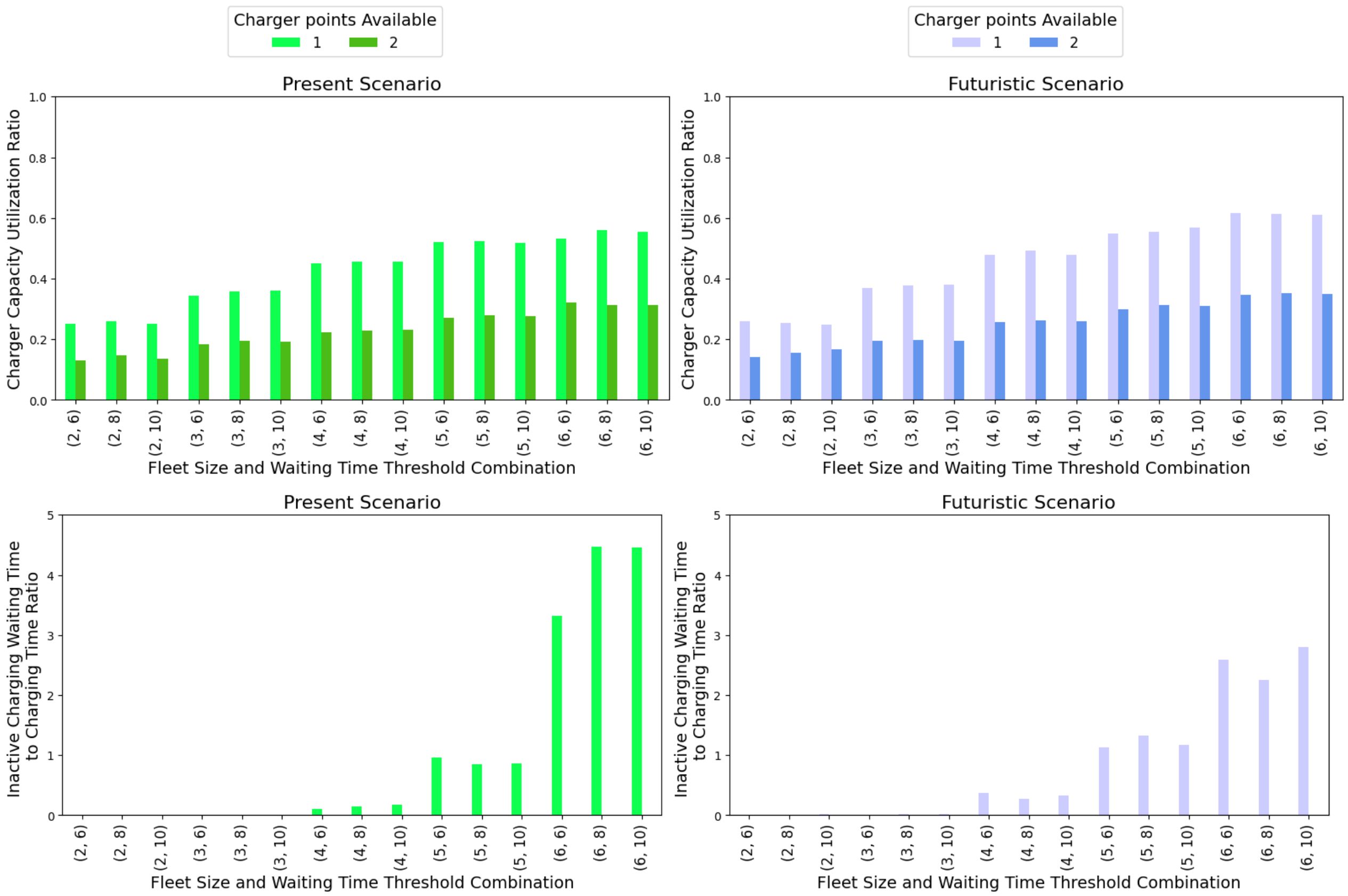

**Figure 12. Charging capacity use ratio and inactive charging waiting time to charging time ratio**

## Traffic Impact Analysis

Traffic impact analysis is performed to check how the shuttle movement in the operating routes affect the regular human driven vehicles since the operating speed of the autonomous shuttle (15 mph) is lower than the regular speed limit (mostly 25 mph) on that road sections. For this experiment, first we run the operator with the busiest configuration (fleet size = 6 and charging points = 2) ensuring the shuttles will be operational with maximum possible time considering the present demand scenario. To analyze the traffic impact, two parameters, speed, and delay time to travel time ratio are used. The parameter speed indicates mean speed of all the human-driven vehicles in all the road sections where shuttles are operating in each 15 minutes interval in mph (see **Figure 13a**). The delay time is the variation of the actual travel time from the ideal travel time (considering no congestion). The delay time to travel time ratio parameter in all the shuttle operating road sections in each 15 minutes interval is shown in **Figure 13b**.

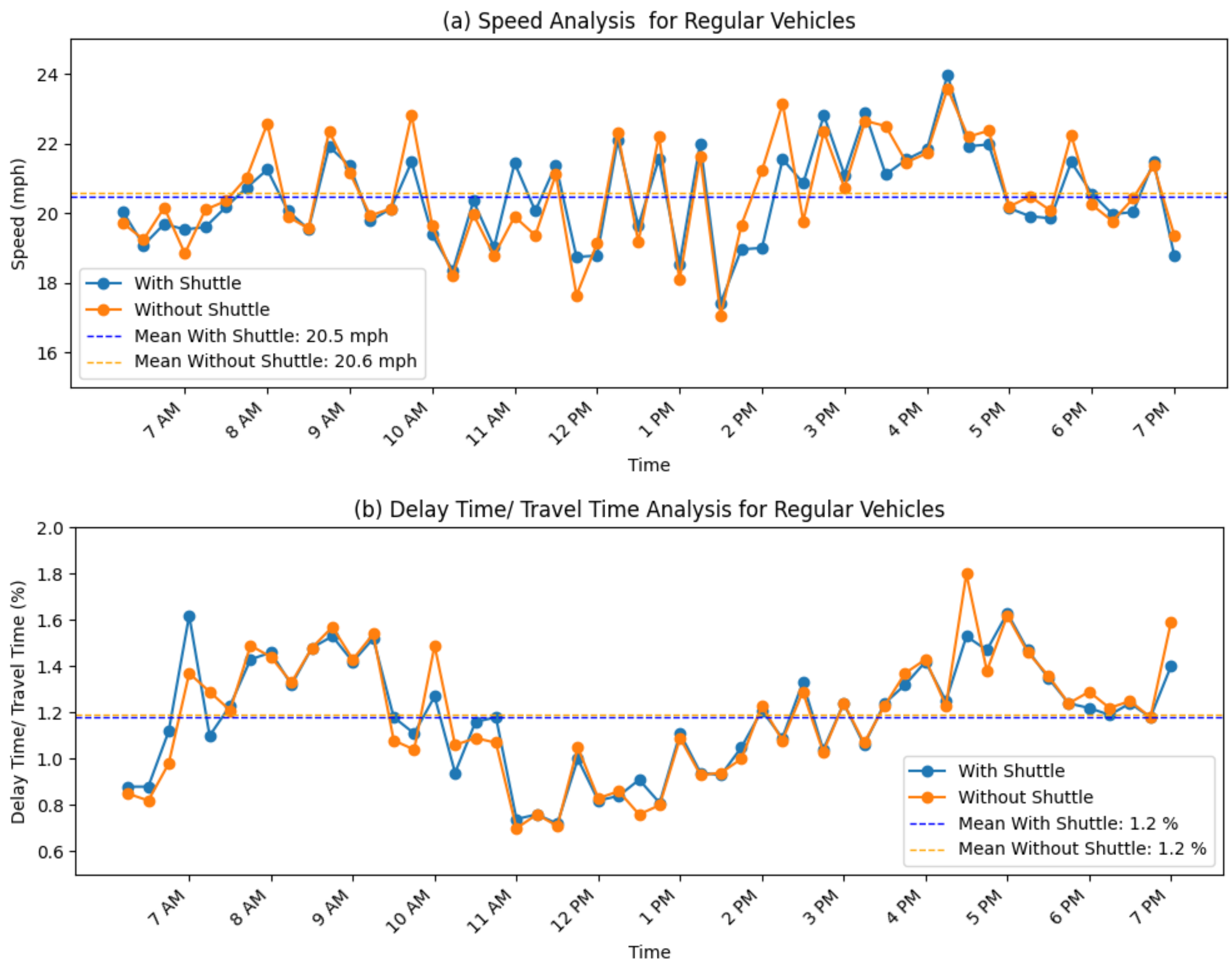

**Figure 13. Comparative analysis of shuttle movement impact on the operating routes**

The results suggest that there will be negligible impact on the movement of regular vehicles by the shuttle operation even with its' busiest configuration. Both the with and without shuttle scenarios are presenting a similar trend for both parameters here, showing very little differences throughout the full simulation time of 13 hours. While the mean speed will decrease in a very negligible amount (0.1 mph) for the shuttle operation, the delay time/travel time ratio will remain largely unchanged. The main reason behind this result is the type of the study area. Since it is a sub-urban area with light traffic demand and vehicle movements, and as the shuttle fleet provides an on-demand service, the regular vehicle movements will not be interfered much by the slow speed of the shuttle.

## DISCUSSION AND CONCLUSIONS

Integrating autonomous electric shuttles for first and last mile (FMLM) connections presents a promising solution to enhance the effectiveness of public transit systems. However, implementing such services requires efficiency analysis from both demand and supply perspectives. The demand-supply interaction is a dynamic process, and agent-based simulations can effectively analyze this mechanism. This research provides a simulation-based methodological framework on how varying operational configurations of an FMLM connector service based on autonomous shuttles can be evaluated. The main contribution of this research is to develop a controller that integrates simulation-based dynamic traffic assignment and heuristics-based optimization for modeling demand-supply interactions. The study provides several insights based on rider, vehicle, energy consumption, and traffic impact perspectives.

One of the key insights from this research is the importance of efficient fleet management. This study shows that a high trip acceptance ratio with minimal waiting times can be achieved through employing a fleet of moderate size of three shuttles. Simulation results suggest that increasing fleet size can improve service levels and reduce waiting times, but there is a point of diminishing returns beyond where additional shuttles do not significantly enhance performance. Empty vehicle miles and shuttle capacity remain stable across different scenarios, which suggests the needs for optimization in vehicle routing and scheduling algorithms and introduction of advanced trip reservation concepts to improve efficiency. The increase in idle time ratios with increasing fleet size underlines the importance of maintaining a balanced fleet size for avoiding under-utilization of the shuttles. The analysis of the charging capacity use ratio reveals the critical role of strategic planning in charging infrastructure. The necessity to consider the waiting time for the chargers is also vital to continue efficient operations considering current and increased future demands. Given the current regulations and state-of-the-art of autonomous vehicle technology, low operating speed of the shuttles can be a drawback for deploying in sub-urban regions. The study shows that low-speed autonomous shuttles have negligible impact on regular vehicle speed and delay time due to shuttle operations.

As such, this study has significant implications for designing transit feeder services. The controller can be a strategic planning tool for autonomous shuttle systems to balance between operational efficiencies, rider satisfaction, environmental benefit, and uninterrupted traffic movement. However, this study has some limitations that can be investigated in future. As the variability in the demand and network infrastructure will influence service performance, diverse urban conditions should be studied. Future works should also consider shuttle relocation, advanced trip reservation, and optimized routing strategies for minimizing overall travel time. There are scopes to investigate different charging strategies like quasi-dynamic wireless charging systems which have minimal effects on shuttle operations. Despite these limitations, the study offers a robust foundation for the future implementation and optimization of autonomous shuttle services in urban transit systems.

## ACKNOWLEDGEMENT

The authors acknowledge Aimsun for providing Aimsun Next and Aimsun Ride license and Teralytics Studio for providing the travel demand data.

ChatGPT was used when preparing the draft of this manuscript to assist in reconstructing or polishing some author-provided text only.

## AUTHOR CONTRIBUTIONS

The authors confirm contribution to the paper as follows: study conception and design: Roy, Nahmias-Biran, Hasan; data collection: Roy; coding: Roy, Dadashev, Yfantis; analysis and interpretation of results: Roy, Nahmias-Biran, Hasan; draft manuscript preparation and editing: Roy, Nahmias-Biran, Hasan, funding and supervision: Hasan. All authors reviewed the results and approved the final version of the manuscript.